\documentclass{bvm} 
\makeatletter
\let\blx@rerun@biber\relax
\makeatother

\begin{document}

\selectlanguage{english} 

\title{Is Self-Supervision Enough? Benchmarking Foundation Models Against End-to-End Training for Mitotic Figure Classification}

\subtitle{}

\titlerunning{Foundation Models for Mitotic Figure Classification}


\author{
	Jonathan \lname{Ganz} \inst{1}, 
	Jonas \lname{Ammeling} \inst{1},
        Emely \lname{Rosbach} \inst{1},
        Ludwig \lname{Lausser} \inst{1},
        Christof~A. \lname {Bertram} \inst{2},
	Katharina \lname{Breininger} \inst{3,4},
	Marc \lname{Aubreville} \inst{5,1}
}



\authorrunning{Ganz et al.}


\institute{
\inst{1} Technische Hochschule Ingolstadt, Ingolstadt, Germany\\
\inst{2} University of Veterinary Medicine, Vienna, Austria \\
\inst{3} Center for AI and Data Science, Julius-Maximilians-Universität Würzburg, Würzburg, Germany\\
\inst{4} Department of Artificial Intelligence in Biomedical Engineering, Friedrich-Alexander-Universität Erlangen-Nürnberg, Erlangen, Germany\\
\inst{5} Flensburg University of Applied Sciences, Flensburg, Germany}

\email{jonathan.ganz@thi.de}
\maketitle


\begin{abstract}
Foundation models (FMs), i.e., models trained on a vast amount of typically unlabeled data, have become popular and available recently for the domain of histopathology. The key idea is to extract semantically rich vectors from any input patch, allowing for the use of simple subsequent classification networks potentially reducing the required amounts of labeled data, and increasing domain robustness.
In this work, we investigate to which degree this also holds for  mitotic figure classification. Utilizing two popular public mitotic figure datasets, we compared linear probing of five publicly available FMs against models trained on ImageNet and a simple ResNet50 end-to-end-trained baseline.
We found that the end-to-end-trained baseline outperformed all FM-based classifiers, regardless of the amount of data provided. Additionally, we did not observe the FM-based classifiers to be more robust against domain shifts, rendering both of the above assumptions incorrect.
 \end{abstract}

\section{Introduction}
Mitotic figures (MFs) in histologic images represent cells undergoing division, and their frequency within a standardized area is a key biomarker for assessing tumor patient survival in both veterinary and human medicine \cite{aubreville2024domain}. However, manual detection of MFs presents a number of challenges, including the tedious nature of the counting process and the low inter-rater agreement \cite{ganz2024information}. Consequently, there is a high level of interest in deep learning-driven detection and classification methods.

Recently, foundation models (FMs) have attracted significant interest in the digital pathology community and there is a rapid development of new models \cite{chen2024uni, vorontsov2024virchow, zimmermann2024virchow2, filiot2023phikon, hoptimus0, xu2024gigapath}. These models, which are typically large vision transformers (ViTs), are trained in an unsupervised manner on extensive collections of histopathology images. As a result of their comprehensive training, they are capable of generating image embeddings with high semantic value, which enables them to perform well in tasks such as few-shot learning and to serve as a robust foundation for a multitude of downstream applications \cite{chen2024uni, vorontsov2024virchow, zimmermann2024virchow2, filiot2023phikon, hoptimus0, xu2024gigapath}.

In recent work, Vonotsov et al.~\cite{vorontsov2024virchow} have shown that the patch embeddings of FMs can be utilized effectively for MF classification. They have reported promising performance by training linear classifiers directly on these patch embeddings, which is of particular interest given that MF classification is often used as a second stage in modern detection pipelines \cite{aubreville2020deep}. As a result, FMs could be employed to further enhance performance of such pipelines or to reduce the need for large amounts of annotated data.
However, there has been no in-depth analysis of how the performance of such classifiers depends on training set sizes or how much it is influenced by domain shifts (i.e., performance degradation due to source-target domain differences). This study aims to shed light on these questions by comparing the embeddings of several state-of-the-art histopathology FMs for MF classification. To provide a comprehensive evaluation, we compare the FM-based classifiers against several baselines across two publicly available MF datasets.
\section{Materials}
In this study, we used two publicly available MF datasets. The first is the training set from the second edition of the Mitosis Domain Generalization Challenge (MIDOG) \cite{aubreville2024domain}, which consists of 11,051 MFs and 9,501 MF look-alikes on 354 regions of interests from five different domains, including human breast cancer (domain 1), canine lung cancer (domain 2), canine lymphoma (domain 3), canine cutaneous mast cell tumor (CCMCT) (domain 4), and human neuroendocrine tumor (domain 5). The second dataset comprises 32 whole slide images (WSIs) of CCMCTs and originally contains 262,481 annotations of different cell types \cite{bertram2019large}. In this study, we only used 44,880 annotations representing MFs and 27,965 annotations belonging to MF look-alikes.
\section{Methods}
We compared six different models in this work, with detailed information provided in Table \ref{tab:models}. All of the models are Vision Transformers (ViTs) of varying sizes. Most of the compared models were trained using DINOv2, a self-supervised learning algorithm that trains vision models by contrasting differently augmented views of the same image, allowing them to learn high-level features without the need for labeled data \cite{oquab2023dinov2}.

\begin{table}[h]
    \centering
    \begin{tabular}{c|c|c|c|c|c}
        Name & Pretraining Algorithm & \multicolumn{2}{c|}{Dataset Size} & Model Type & Model Size \\
        & & WSI & Tiles & & \\
        \hline
        Prov-GigaPath \cite{xu2024gigapath} & DINOv2 &170K &1.3B &ViT-G &1.1B \\
        H-optimus-0 \cite{hoptimus0} & DINOv2 &500K & - &ViT-G &1.1B \\
        Virchow \cite{vorontsov2024virchow} & DINOv2 & 1.5M & 2B &ViT-H &632M \\
        Virchow2 \cite{zimmermann2024virchow2} & DINOv2 &3.4M &1.7B &ViT-H &632M \\
        UNI \cite{chen2024uni}& DINOv2 & 100K & 100M& ViT-L& 307M\\
        Phikon \cite{filiot2023phikon} & iBOT &6K &43M &ViT-B &86M \\
    \end{tabular}
    \caption{Foundation models compared in this study, ordered by the size of the respective model.}
    \label{tab:models}
\end{table}

For MF classification, we used the linear probing implementation by Chen et al.~\cite{chen2024uni}, which trains a linear classifier on top of pre-extracted embeddings. To generate embeddings, we extracted patches of size $224\times224$ pixels for the MFs and look-alikes at a resolution of 0.25 microns per pixel. For Virchow and Virchow2, these embeddings were created by concatenating the class token and the mean across all 256 patch tokens, as described in \cite{vorontsov2024virchow}. For all other models, only the class token was used \cite{vorontsov2024virchow}. During image extraction, the patches were normalized according to the means and standard deviations provided in the respective works. No additional data augmentation was applied.

We compared the embeddings of the FMs to three different baselines. First, we used two feature extractors (ResNet50 with 25M and ViT-H 632M parameters) pretrained on ImageNet. As with the FMs, these models were used to generate embeddings for linear probing.
Additionally, we compared the models to an end-to-end-trained baseline. For this, we used a classifier based on a ResNet50 stem pretrained on ImageNet. Unlike the embedding generation process, we applied a standard image augmentation pipeline during the training, which included random color jitter, Gaussian blurring, flipping, and random rotations. Besides this, the patches were extracted as for the FMs. We used the one-cycle learning rate policy with a maximum learning rate of $10^{-4}$ \cite{smith2017super}, and the best-performing model was selected retrospectively based on its validation loss.

To examine performance as a function of training set size, we trained the models using different subsets of the two datasets. To increase the statistical informativeness of our results, we applied five-fold Monte Carlo cross-validation for each model and dataset size. With four different training set sizes ($0.1\%$, $1\%$, $10\%$, and $100\%$) a total of 20 trainings were performed per model.  During each training, $20\%$ of all annotations of the respective dataset were randomly selected as test set, while $20\%$ of the respective training annotations were selected for validation. To ensure comparability, all models were trained and evaluated using the same training and test sets. Please note that the percentages refer to the size of each dataset, meaning substantially less data was available for the respective training set sizes in the MIDOG dataset compared to the CCMCT dataset. In a second experiment to assess cross-domain performance, we split the MIDOG dataset into training and test sets across different domains. For each evaluation, one domain served as the training set, while the remaining domains were used for testing. To assess in-domain performance, 20\% of the in-domain data was separated from the training set and included in the test set. At the end of the experiment, performance was measured separately for each domain. As in the previous experiment, five training sessions were conducted per domain and per model, resulting in a total of 25 training sessions per model. 

\section{Results}
\begin{figure}[h]
    \centering
    \includegraphics[width=1\linewidth]{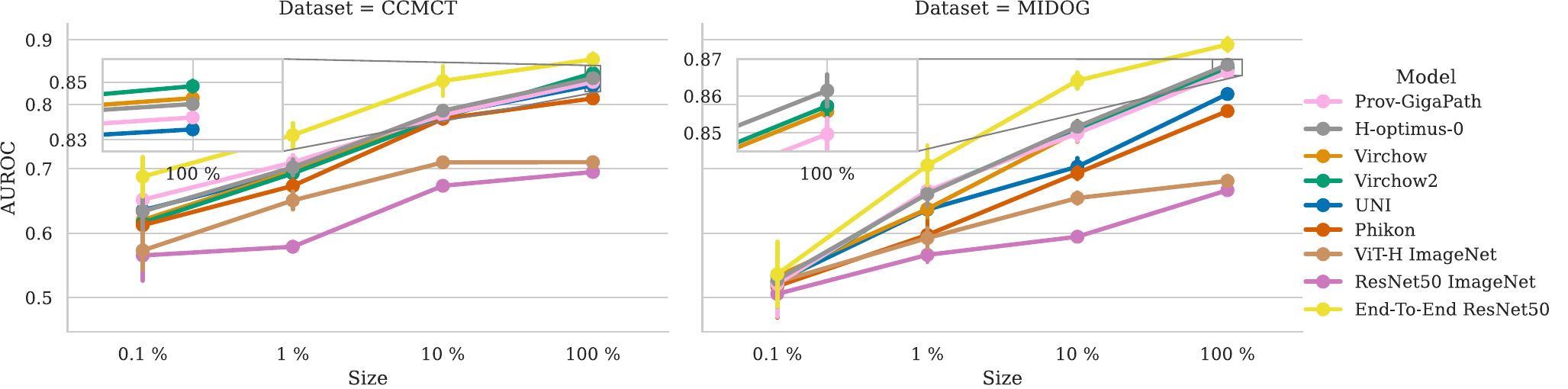}
    \caption{Results of the investigation of the influence of training set size on model performance. The percentages refer to the total size of the data set.}
    \label{fig:data_set_size}
\end{figure}

The results of the analysis on the influence of dataset size on model performance are shown in Figure \ref{fig:data_set_size}. We observed that, for all models, performance improved significantly as the size of the available training data increased. On the CCMCT dataset, the performance increase in relation to the amount of training data was similar across all FM-based classifiers. However, on the MIDOG dataset, classifiers using embeddings from Phikon and UNI lagged behind the other FM-based approaches. The two ImageNet baselines were significantly outperformed by both the end-to-end trained ResNet50 and the FM-based classifiers on both datasets. Across data set sizes, the end-to-end trained baseline model outperformed all other models across both datasets. With $100\%$ of the respective dataset available, it achieved a mean AUROC of $0.87\pm 0.01$ on the CCMCT dataset and $0.89\pm 0.01$ on the MIDOG dataset. Among the classifiers based on FM embeddings, the H-optimus-0, Prov-GigaPath, Virchow, and Virchow2 models demonstrated similar performance. With $100\%$ of the data available, on the CCMCT dataset, Virchow2 embeddings had a slight advantage over the other FMs, while on the MIDOG dataset, H-optimus-0 embeddings delivered the best performance among the FMs.

\begin{figure}[h]
    \centering
    \includegraphics[width=0.82\linewidth]{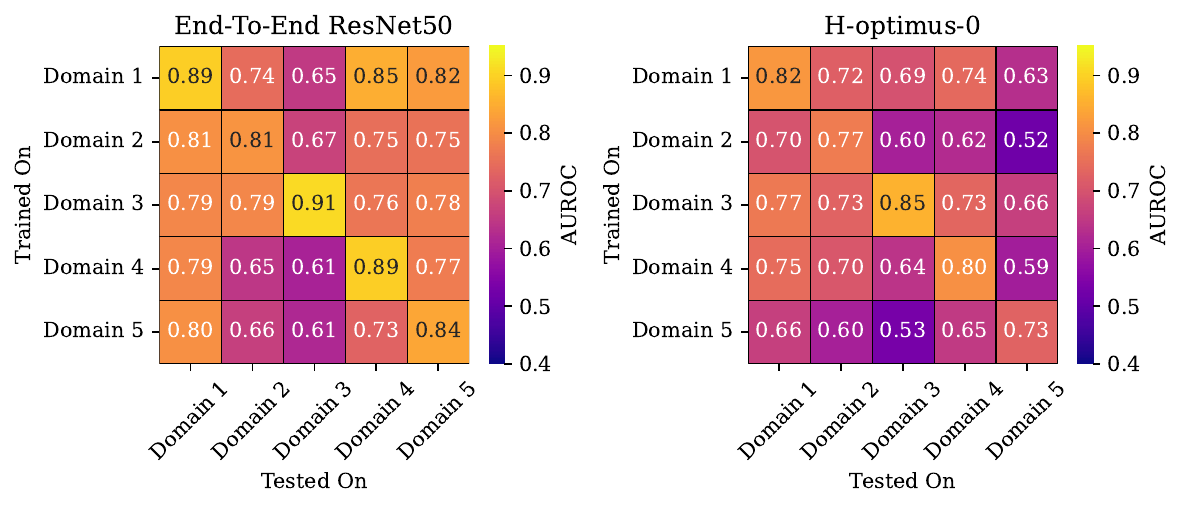}
    \caption{Cross-domain AUROC results for the end-to-end trained ResNet50 baseline model and the classifier based on H-optimus-0 embeddings.}
    \label{fig:domain}
\end{figure}

The mean in-domain and out-of-domain performances over the different domains for all models are given in Table \ref{tab:domain}. As embeddings of the the H-optimus-0 model gave the best results in the first experiment, we choose to compare it to the end-to-end baseline in more detail in Figure \ref{fig:domain}.

Like in the first experiment, the end-to-end baseline clearly outperforms all other models, both in terms of in-domain performance but also in terms of out-of-domain performance (see Table \ref{tab:domain}). Among the FMs, the classifiers using the embeddings of the H-optimus-0, Prov-GigaPath, Virchow, and Virchow2 models performed slightly better then those using the UNI and Phikon embeddings. However, given the standard deviations, the performance were relatively similar.

Both the FM-based classifiers and the end-to-end baseline experienced substantial performance drops when the in-domain trained models were applied to different domains, as shown in Table \ref{tab:domain}. Additionally, Figure \ref{fig:domain} demonstrates that, for both the end-to-end trained ResNet50 and the H-optimus-0 models, this drop in performance was not evenly distributed across the external domains. This observation aligns with the findings of \cite{aubreville2024domain}, which highlight that domain shift is not a symmetrical phenomena.

\begin{table}[]
\caption{In-domain and out-of-domain AUROC results for different models on the MIDOG dataset (mean $\pm$ standard deviation of five-fold Monte Carlo cross-validation). }
\label{tab:domain}
\begin{tabular}{c|c|c}
Model               & AUROC In Domain & AUROC Out of Domain \\ \hline
Prov-GigaPath       & 0.79    $\pm$  0.04             &0.66                  $\pm$ 0.07                \\
H-optimus-0         & 0.79               $\pm$ 0.04             &0.66                  $\pm$ 0.07               \\
Virchow             & 0.79               $\pm$ 0.04             &0.66                  $\pm$ 0.06                \\
Virchow2            & 0.79               $\pm$ 0.04             &0.66                  $\pm$ 0.06                \\
UNI                 & 0.76               $\pm$ 0.03             &0.64                  $\pm$ 0.05                \\
Phikon              & 0.76               $\pm$ 0.03             &0.61                  $\pm$ 0.05                \\
ViT-H ImageNet      & 0.68               $\pm$ 0.05             &0.56                  $\pm$ 0.04                \\
ResNet50 ImageNet   & 0.63               $\pm$ 0.04             &0.53                  $\pm$ 0.03                \\
End-To-End ResNet50 & \textbf{0.87}                  $\pm$ 0.04             &\textbf{0.74}                  $\pm$ 0.07               
\end{tabular}
\end{table}

\section{Discussion}
In this study, we compared classifiers based on FM embeddings to different baseline models. In both experiments, the end-to-end-trained ResNet50 baseline outperformed the FM-based classifiers. Thus we conclude that there is no benefit in employing the embeddings generated by those models directly for MF classification. 
It should be noted that the FM-based classifiers consisted of only a single linear layer trained with pre-extracted features, whereas the end-to-end model was fully trainable. Nevertheless, the FMs underwent extensive pretraining on large histopathology datasets, while the ResNet baseline, despite being pretrained on ImageNet, was only trained on the data available for each experiment. This suggests that extensive pretraining is no substitute for task-specific supervision, at least for the given task. However, better results might be achievable if the FMs were fine-tuned for the specific task, though this would require significantly more computational resources and might be inefficient given the strong performance of the standard CNN baseline.
The positive impact of the FMs' pretraining- on the given tasks suggests a comparison with the ViT-H based classifier, which, despite having a comparable number of parameters, achieved lower performance (see Fig. \ref{fig:data_set_size}).

The FM-based classifiers didn't turn out to be more data-efficient in comparison to our end-to-end trained ResNet50 model. However, the performance on both datasets in Fig. \ref{fig:data_set_size} does not seem to have converged yet. Hence, it might be interesting to see how the performance scales with even more data being available.


We also did not find the FM-based classifiers to be more robust to domain shifts than one of our baselines. This suggests that the pretraining of the FMs did not result in entirely domain-agnostic embeddings. Interestingly, the in-domain results of the FM-based classifiers in Table \ref{tab:domain} show that all models performed quite similarly, whereas in the first experiment, the Uni and Phikon-based models performed slightly worse on the MIDOG dataset compared to the other FMs. Indeed, the results in Table \ref{tab:domain} resemble those on the CCMCT dataset, where all FM-based classifiers performed similarly. This may indicate that the Uni and Phikon embeddings are marginally less generalizable than those of the other FMs, as the latter were able to perform better when trained on the mixed MIDOG domains in the first experiment.

\begin{acknowledgement}
M.A. and C.B. acknowledge support by the DFG and FWF, respectively (project numbers 520330054, I 6555). K.B. acknowledges support of the DFG, project 460333672 CRC1540 EBM.
\end{acknowledgement}

\printbibliography
\end{document}